\pdfoutput=1

\documentclass[11pt]{article}

\usepackage[preprint]{acl}

\usepackage{times}
\usepackage{latexsym}
\usepackage{graphicx}  
\usepackage{subcaption} 
\usepackage{caption}    
\usepackage{booktabs}
\usepackage{multirow}
\usepackage{float}
\usepackage{listings}
\usepackage{array}
\usepackage{makecell}
\usepackage{pifont}

\usepackage[T1]{fontenc}

\usepackage[utf8]{inputenc}
\usepackage{stfloats}

\usepackage{microtype}

\usepackage{inconsolata}

\usepackage{graphicx}
\usepackage{tcolorbox}
\usepackage{xcolor}

\definecolor{rq1}{RGB}{177, 206, 231}  
\definecolor{rq2}{RGB}{248, 180, 196}  
\definecolor{rq3}{RGB}{255, 231, 173}  
\definecolor{rq4}{RGB}{156, 252, 227}  

\usepackage{enumitem}
\usepackage{verbatim}
\usepackage{pdfpages}
\usepackage{hyperref}

\title{\textsc{CaMMT}: Benchmarking Culturally Aware \\ Multimodal Machine Translation}

\author{
 \textbf{Emilio Villa-Cueva\textsuperscript{\ding{61},\ding{72}}},
 \textbf{Sholpan Bolatzhanova\textsuperscript{\ding{61},\ding{72}}},
 \textbf{Diana Turmakhan\textsuperscript{\ding{61},\ding{72}}},
 \textbf{Kareem Elzeky\textsuperscript{\ding{72}}},
\\
 \textbf{Henok Biadglign Ademtew},
 \textbf{Alham Fikri Aji},
 \textbf{Vladimir Araujo},
 \textbf{Israel Abebe Azime},
\\
\textbf{Jinheon Baek},
 \textbf{Frederico Belcavello},
 \textbf{Fermin Cristobal},
 \textbf{Jan Christian Blaise Cruz},
\\
\textbf{Mary Dabre},
 \textbf{Raj Dabre},
 \textbf{Toqeer Ehsan},
 \textbf{Naome A Etori},
 \textbf{Fauzan Farooqui},
\\
 \textbf{Jiahui Geng},
 \textbf{Guido Ivetta},
 \textbf{Thanmay Jayakumar},
 \textbf{Soyeong Jeong},
\\
 \textbf{Zheng Wei Lim},
 \textbf{Aishik Mandal},
 \textbf{Sofía Martinelli},
 \textbf{Mihail Minkov Mihaylov},
\\
 \textbf{Daniil Orel},
 \textbf{Aniket Pramanick},
 \textbf{Sukannya Purkayastha},
 \textbf{Israfel Salazar},
\\
 \textbf{Haiyue Song},
 \textbf{Tiago Timponi Torrent},
 \textbf{Debela Desalegn Yadeta},
 \\
 \textbf{Injy Hamed\textsuperscript{\ding{72}}},
 \textbf{Atnafu Lambebo Tonja\textsuperscript{\ding{72}}},
 \textbf{Thamar Solorio\textsuperscript{\ding{72}}}
\\
 \small{\textbf{\textsuperscript{\ding{72}} Core Authors (MBZUAI)}}
}

\begin{document}
\maketitle

\begin{abstract}
\let\thefootnote\relax\footnotetext{\ding{61} Equal Contribution}
Translating cultural content poses challenges for machine translation systems due to the differences in conceptualizations between cultures, where language alone may fail to convey sufficient context to capture region-specific meanings. In this work, we investigate whether images can act as cultural context in multimodal translation. We introduce \textsc{CaMMT}, a human-curated benchmark of over 5,800 triples of images along with parallel captions in English and regional languages. Using this dataset, we evaluate five Vision Language Models (VLMs) in text-only and text\textbf{+}image settings. Through automatic and human evaluations, we find that visual context generally improves translation quality, especially in handling Culturally-Specific Items (CSIs), disambiguation, and correct gender marking. 
By releasing \textsc{CaMMT}\footnote{\url{https://huggingface.co/datasets/villacu/cammt}}, our objective is to support broader efforts to build and evaluate multimodal translation systems that are better aligned with cultural nuance and regional variations.

\end{abstract}
\begin{figure*}[b]
    \centering    \includegraphics[width=\textwidth]{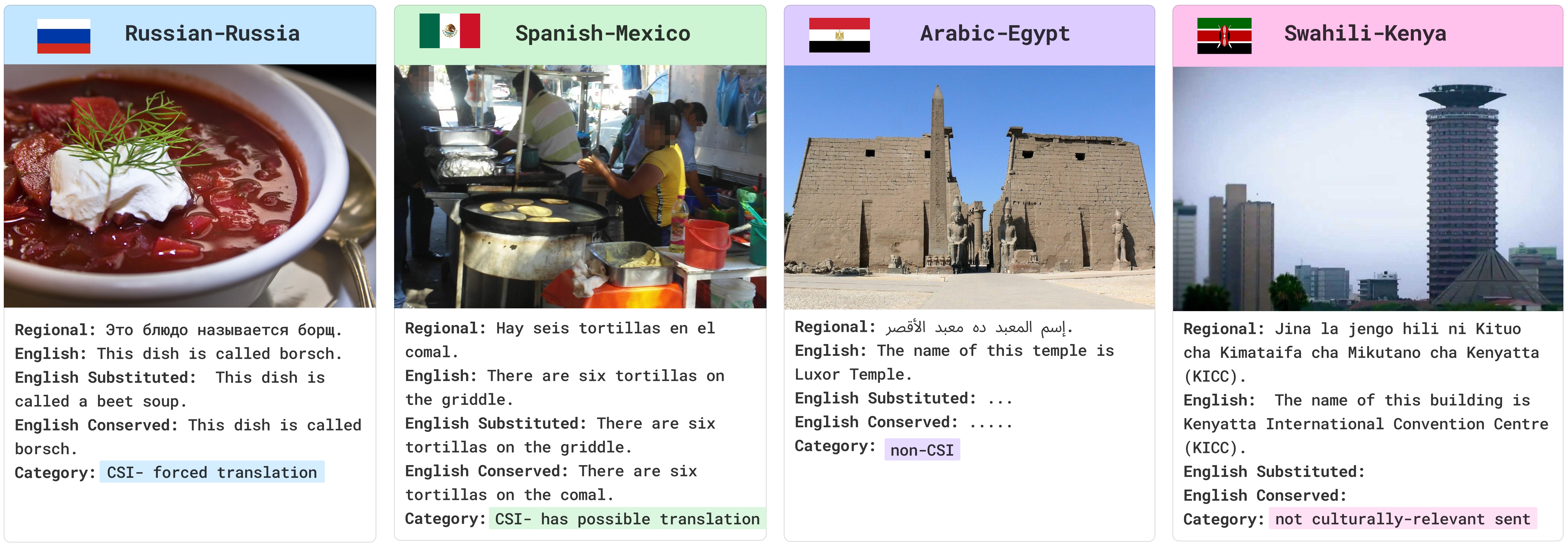}
    \caption{Examples of \textsc{CaMMT} dataset}
    \label{fig:cvqammt_example}
\end{figure*}

\section{Introduction}
Translation brings cultures into contact. It usually involves deciding how much foreignness to keep in the resulting translation and invariably involves blending cultures to some extent \cite{axiela}. As pointed out by \citet{hershcovich2022challenges}, part of the difficulty in deciding the right level of culture blending during translation arises from the different conceptualizations that each culture holds. Translators must, therefore, choose suitable strategies for adapting vocabulary as well as deciding whether to conserve or substitute foreign elements. 
Conforming the source text to the target culture by substituting unknown elements with familiar ones can ease comprehension, yet it simultaneously erases traces of the original culture \cite{venuti}. Conversely, ignoring an adequate vocabulary choice that accounts for regional variation in the target language risks misinterpretation, as lexical choice directly shapes how readers understand a text \cite{Szymaska2017}.

Text-only machine translation inherits this dilemma with limited contextual knowledge to ground these translation decisions. However, images can supply that missing extra-linguistic information; visual reference may act as a cultural proxy, revealing a region's set of values \cite{yadav2025wordsexploringculturalvalue} as well as social practices and material culture, such as clothing, architecture, and food. With photography being thought of as a form of translation from reality into images \cite{gagliano2008photography}, we hypothesize that images can capture additional information that language alone may struggle to encode.

Multimodal Machine Translation (MMT) \cite{specia2016shared} attempts to embed this information by grounding source sentences with images. CoMMuTe \cite{futeral2022tackling} provides an evaluation framework for MMT centered on lexical disambiguation, but does not address broader cultural nuances, leaving questions about how visuals influence translation in culturally grounded settings largely unanswered.

In this work, we present \textsc{CaMMT} ( \textbf{C}ulturally-\textbf{A}ware \textbf{M}ultimodal \textbf{M}achine \textbf{T}ranslation Benchmark), the first human-curated MMT corpus with triples across 19 languages of culture-related captions spanning 23 regions worldwide. Additionally, we study the impact of visual grounding for culture-aware multimodal machine translation in Vision–Language Models (VLMs).

To frame our study, we pose the following \textbf{research questions}:
    \begin{itemize}
        \item \colorbox{rq1}{\textbf{RQ1}}: How does visual grounding impact translation quality and native speakers' preferences across different languages in culturally-relevant settings?
        \item \colorbox{rq2}{\textbf{RQ2}}: What reasons drive preferences between text-only and multimodal translations?
        \item  \colorbox{rq3}{\textbf{RQ3}}: How do VLMs perform in MT compared to each other and to state-of-the-art machine translation models?
        \item  \colorbox{rq4}{\textbf{RQ4}}: Which translation strategies do native speakers prefer in the case of CSIs? 
    \end{itemize}
Our contributions are as follows:
    \begin{itemize}

        \item \textbf{Culturally-Specific MMT Dataset}: We present \textsc{CaMMT}, a human-curated corpus of $5,817$ image-captions triples, where the captions are collected for both English and regional languages. For triples containing CSIs, we also provide a separate split with 1,550 samples, where each includes two English translations: one conserving the term and another substituting it.
        \item \textbf{Insights into visual grounding for culture-aware translation}: We evaluate five VLMs on \textsc{CaMMT} to assess the impact of visual grounding on human preferences and performance in automatic metrics. Through these experiments, we find that visual context improves translation outputs. Native speakers tend to prefer multimodal translations because they better preserve CSIs, resolve lexical ambiguities, and reflect proper gender marking, highlighting aspects of translation quality ignored by standard evaluation metrics.
        
    \end{itemize}

\section{Related Work}
\label{sec:related_work}

In translation studies, CSIs \cite{axiela} refer to words or concepts that lack direct equivalents or carry different connotations in the target culture. These often arise when cultural references embedded in the source language do not directly exist or are understood differently in the target language. When translating CSIs, translators typically adopt one of two strategies: \textit{substitution}, which adapts the foreign element into a culturally familiar counterpart to reduce its strangeness; or \textit{conservation}, which preserves the original cultural reference, maintaining the source text's foreignness and exposing readers to its original context \cite{axiela, venuti}.

Efforts to incorporate cultural awareness into machine translation have been addressed in specific domains such as cultural adaptation in recipe translation \cite{cao2024cultural, zhang2024cultural}.  \citet{yao2023benchmarking} generalized beyond this scope by constructing an evaluation dataset by automatically extracting CSIs from Wikipedia to study how LLMs and MT systems handle cultural references. However, the dataset is restricted to a smaller number of languages, automatically generated without input from regional speakers, and does not consider the effect of visual context on translation decisions.

Recent benchmarks such as CVQA \cite{romero2024cvqa}, CulturalVQA \cite{nayak2024benchmarkingvisionlanguagemodels}, ALM‑bench \cite{vayani2025languagesmatterevaluatinglmms}, and FoodieQA \cite{li2024foodieqa} demonstrate growing progress in regional image understanding within VLMs. However, none of these works study how imagery can affect translation across cultures. Together, these studies motivate our evaluation on the multimodal translation ability of VLMs.

\begin{figure*}[t]
    \centering    
    \includegraphics[width=\textwidth]{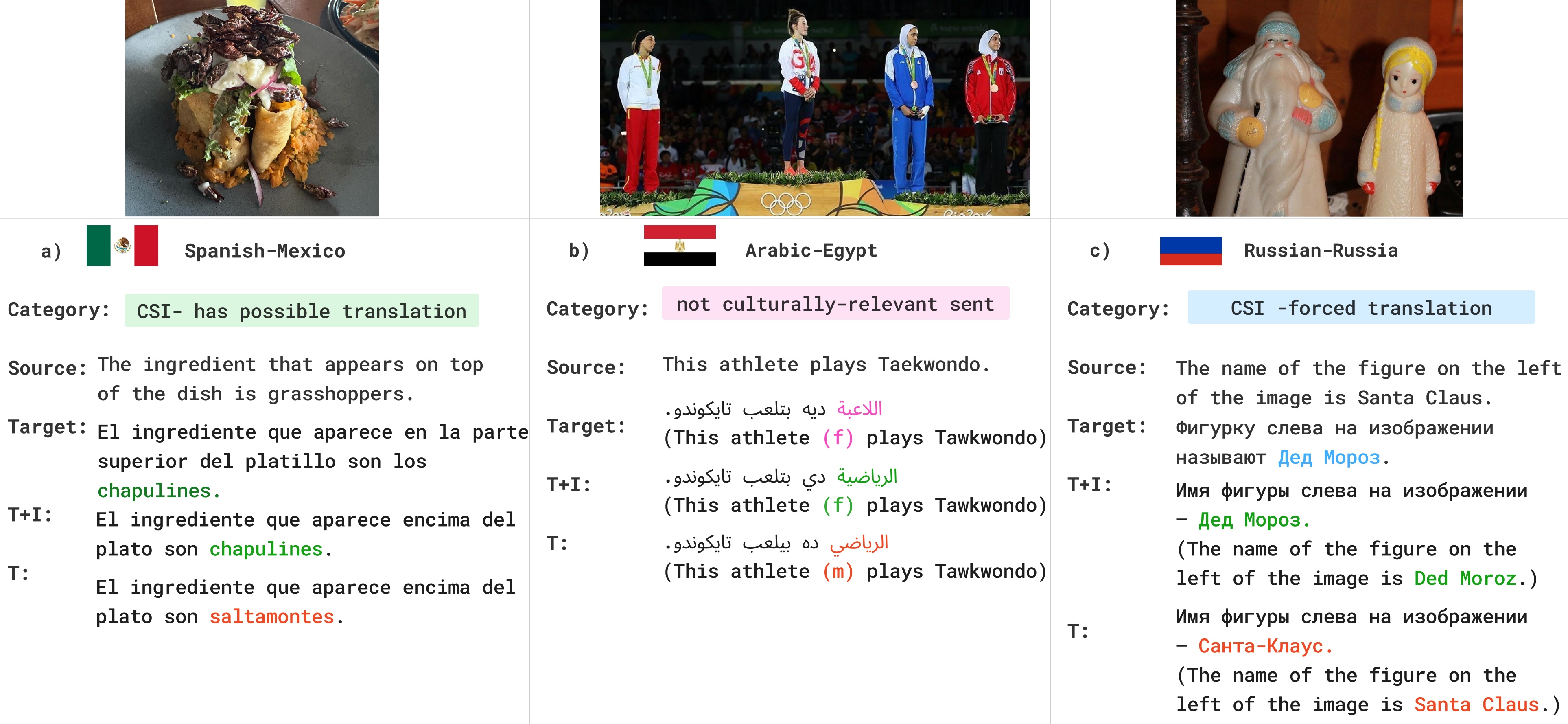}
    \caption{Examples where the text\textbf{+}image translation was marked as preferred over the text-only setting. Image (a) is generated by Gemma3 27B, while (b) and (c) are from Qwen2.5-VL 32B. Examples (a) and (c) illustrate translations preferred because of CSI-preservation, while (c) was preferred as the correct gender of ``athlete'' was used when translating from English to Arabic (a gender-marking language).}
    \label{fig:example}
\end{figure*}

\section{\textsc{CaMMT} Dataset}

CVQA \cite{romero2024cvqa} is a visual question answering dataset comprising more than 10,000 questions across 39 country-language pairs. The questions within CVQA are formulated in both regional languages and English, classified into 10 distinct categories.  
To develop \textsc{CaMMT}, we utilized CVQA's question-answer pairs and transformed them into declarative statements using Gemini 2.0 Flash \cite{geminiteam2024geminifamilyhighlycapable} to generate parallel caption pairs in English and regional languages. 

No images were used in this process to ensure that the phrasing of these seed captions (later refined by annotators) was not influenced by them. The simplicity of the statements, combined with human curation, further reduced the risk of any bias from the language model.

\paragraph{Human Annotations} To ensure the correctness of the generated caption pairs, we involved native speakers (annotators) for each of the languages that participated in the original data curation and are co-authors of this paper.  
The annotators were asked to complete three tasks: \textbf{(1)} evaluate and ensure the grammatical correctness and parallelism of the generated pairs in English and regional language by correcting captions when needed, \textbf{(2)} ensure CSIs in regional language captions are preserved and \textbf{(3)} categorize each of the pairs into three categories: \textbf{(a)} Not culturally relevant sentences, \textbf{(b)} culturally relevant, but do not contain any CSI (\textit{Non-CSI}) or \textbf{(c)} contain CSI. 

We borrowed the definition of CSIs provided to annotators from \citet{axiela}. To achieve a better coverage of translation strategies for CSIs (as previously discussed in Section \ref{sec:related_work}), we asked them to further categorize sentence pairs marked as containing CSIs into \textbf{(i)} CSI with possible translation - captions containing CSIs that have culturally equivalent terms that can convey an equivalent meaning when translated into English and \textbf{(ii)} CSI forced translation - captions containing CSIs that do not have any equivalent translation in English.  For each sentence containing CSIs, we asked the annotators to provide both \textit{conserved} (retaining CSIs) and \textit{substituted} (using familiar equivalents) English translations, then select their preferred version as native speakers. 

For example, in the possible translation category, the Mexican term \textit{tianguis} can be translated as flea market, as in: “The name for this type of Mexican informal market is \textit{tianguis}” (\textit{conserved}) or “The name for this type of Mexican informal market is flea market” (\textit{substituted}). In contrast, a forced translation case is: “The name of the Egyptian food in the glass plate in the picture is \textit{Hawawshi}” (\textit{conserved}) and “The name of the Egyptian food in the glass plate in the picture is \textit{minced meat sandwich}” (\textit{substituted}), where the original term lacks an exact English equivalent. For forced translations, they provide the closest possible English approximation. We provide the annotation guidelines in Appendix~\ref{sec:annotation_guideline}.

\paragraph{Dataset Statistics} In total, \textsc{CaMMT} comprises 23 regions with 19 different languages, with a total of 5,817 triples with additional 1,550 with \textit{conserved} and \textit{substituted} CSIs for targeted analysis. We present representative samples in Figure \ref{fig:cvqammt_example}, and report the number of triples per language included in the corpus in Appendix \ref{sec:data_statistics}.

\section{VLMs for Multimodal Machine Translation}
\label{sec:vlms}

This section explains our motivation for using VLMs as off-the-shelf MMT systems and our evaluation framework. We first present the selected models and validate their effectiveness for the task, followed by a description of our evaluation setup, which measures translation quality through human and automatic assessments in both text-only and text+image conditions in \textsc{CaMMT}.


As discussed in Section \ref{sec:related_work}, task-specific MMT models are limited by their training data, often lacking coverage for many languages. On the other hand, LLMs have demonstrated strong performance in machine translation 
across multiple language pairs \cite{hendy2023goodgptmodelsmachine,zhu2024multilingualmachinetranslationlarge}. As the paradigm shifts from text-only to multimodal LLMs which can process both text and images (VLMs), we explore their potential for multimodal translation, particularly in culturally grounded scenarios.

\begin{table}[h]
\centering
\resizebox{\columnwidth}{!}{%
\begin{tabular}{lllll}
\toprule
Model & Setting & De & Fr & Ru \\ 
\midrule
mBART+MT & T & 25.9 & 38.2 &  \\
VGAMT & T+I & 29.3 (+3.4) & 32.2 (-2.3) &  \\
NLLB-600M & T & 36.2 & 39 & 19.4 \\
NLLB-3.3B & T & 40.8 & 41.4 & 23.1 \\ \hline
Gemma3 27B & T & 39.1 & 41.7 & 23.2 \\
Gemma3 27B & T+I & 44.9 (\textbf{+5.8}) & 49.6 (\textbf{+7.9}) & 31.7 (\textbf{+8.4}) \\
Qwen2.5 VL 32B & T & 32.8 & 33.1 & 21.4 \\
Qwen2.5 VL 32B & T+I & 37.0 (\textbf{+4.2}) & 41.7 (\textbf{+8.6}) & 24.1 (\textbf{+2.7}) \\
Gemini 2.0 Flash & T & 42.6 & 43.1 & 26.8 \\
Gemini 2.0 Flash & T+I & 49.9 (\textbf{+7.3}) & 55.2 (\textbf{+12.1}) & 32.3 (\textbf{+5.5}) \\ 
\bottomrule
\end{tabular}%
}
\caption{
BLEU scores reported on CoMMuTe for text-only (T) and text\textbf{+}image (T+I) settings. The scores from mBART+MT and VGAMT (an MMT system based on BART) are as reported by \citet{futeral2022tackling}, who does not evaluate Russian.}
\label{tab:commute}
\end{table}

To initially assess the ability of VLMs in grounding translations using images, we conduct a control experiment on the CoMMuTe dataset \cite{futeral2022tackling}, comparing them against strong task-specific MT and MMT baselines. CoMMuTe consists of English sentences with ambiguous terms paired with two images that lead to different translations (e.g., \textit{mole} may refer to an animal or a skin mark). Thus, improvements when images are provided indicate that the model is effectively leveraging visual context to disambiguate the source sentence. 

In the text-only setting, models are prompted to translate from English to the target language. In the text\textbf{+}image setting, they are additionally provided with an image and prompted to use it as context for the translation (see Appendix~\ref{sec:translation_prompts} for prompt details). Importantly, no further instructions are given regarding the nature of the disambiguation task. We evaluate five VLMs: Gemma 3 27B and 12B \cite{gemmateam2025gemma3technicalreport}, Qwen 2.5-VL 32B and 8B \cite{bai2025qwen25vltechnicalreport}, and Gemini 2.0 Flash \cite{geminiteam2024geminifamilyhighlycapable}.

Results presented in Table~\ref{tab:commute} demonstrate consistent and significant improvements in the performance of VLMs in the text\textbf{+}image over text-only setting. Moreover, the BLEU scores achieved by VLMs match or surpass those of NLLB-600M and NLLB-3.3B, strong baselines, as well as dedicated MMT systems. These results confirm that VLMs can indeed leverage visual context to guide translation decisions. Based on this validation, we continue with VLMs as our testbeds to probe how visual grounding influences translation choices in our culturally relevant dataset. 




We focus on the Gemini, Gemma, and Qwen families, covering closed-weight and open-weight models at different scales. For completeness, we also report automatic evaluation results for Aya-Vision \cite{cohereVisionExpanding} in Appendix~\ref{sec:aya_appendix}, consistent with the main models' findings.

All models are evaluated in both text-only and text\textbf{+}image setups. In the text\textbf{+}image setting, we do not explicitly instruct models to use images as a cultural reference, only as additional context, allowing us to observe their default effect in translation. 
To evaluate the impact of visual input on translation quality, we conduct both \textit{human preference evaluation} and \textit{automatic evaluation} using standard machine translation metrics and the curated pairs as ground truth.

\begin{figure*}[t!]
    \centering
    \includegraphics[width=1\linewidth]{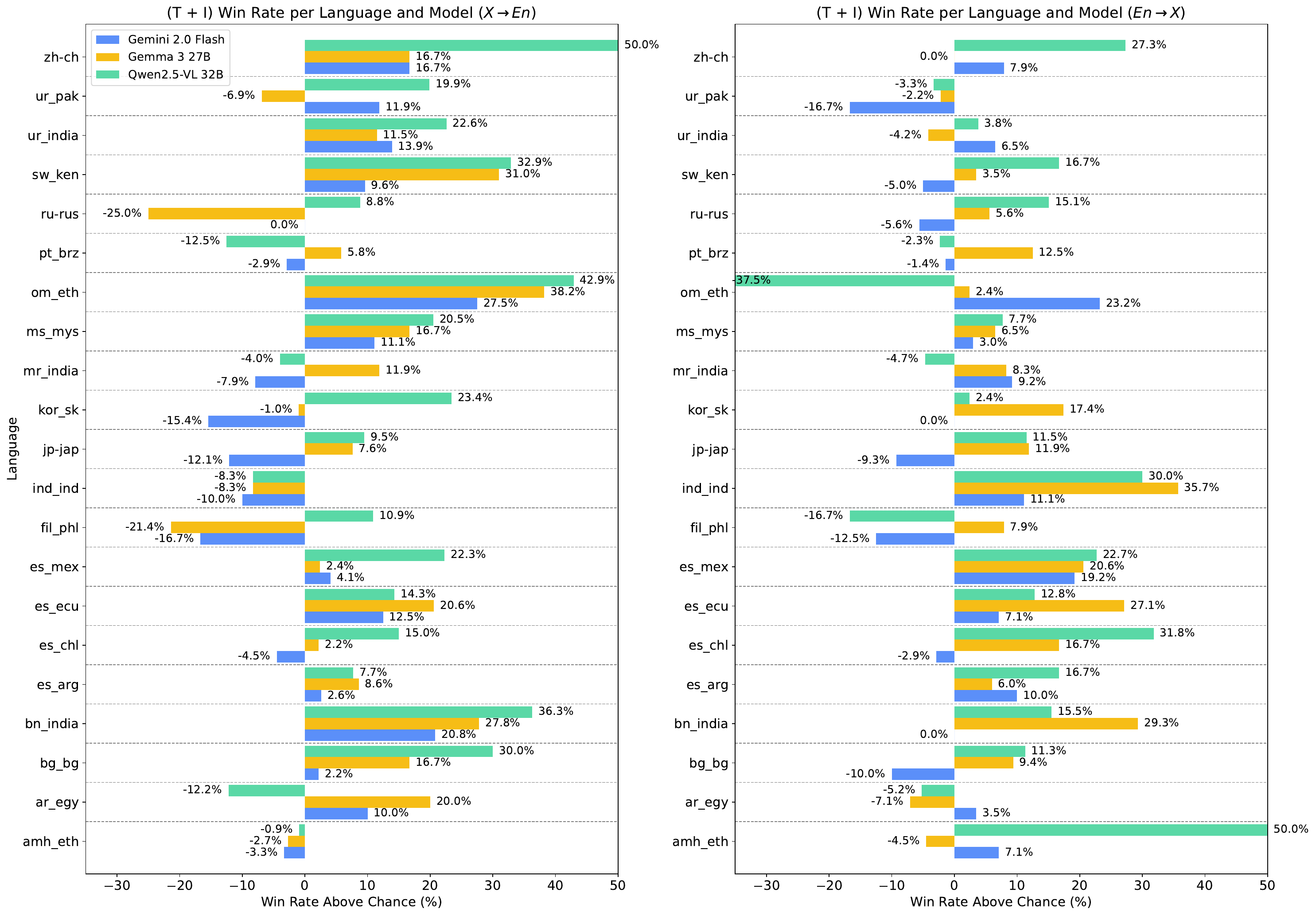}
    \caption{Win rates in human preference evaluation of text\textbf{+}image (T+I) translations over text-only (T) across languages and models. Each bar represents the win rate \textit{above chance} (i.e., over 50\%) for cases where native speakers expressed a preference between the two translation conditions. The left plot corresponds to the $X \to En$ direction, and the right to $En \to X$.}
    \label{fig:mmt-win-rates}
\end{figure*}

\paragraph{Human Preference Evaluation Setup} 

For 21 of the \textsc{CaMMT} regions, native speakers are presented with anonymized translations from three models—Qwen2.5-VL 32B, Gemma 3 27B, and Gemini 2.0 Flash—generated under both text-only and text\textbf{+}image settings. For each instance, they select the preferred translation and specify the reason for their preference from a predefined set: “\textit{CSI is preserved},” “\textit{Correct gender},” “\textit{Disambiguates word},” or “\textit{Regionally appropriate phrasing}”. We identified this set of reasons based on an analysis carried out in preliminary experiments on a subset of languages. Annotators are also allowed to specify \textit{other} reasons if none of the previous reasons explain the preference. In Appendix~\ref{sec:preference_guidelines}, we present the instructions provided for this evaluation.

\paragraph{Automatic Evaluation Setup} We automatically evaluate translation quality using BLEU \cite{PRT+02}, chrF++ \cite{popovic2017chrf++}, and BERTScore (F1) \cite{ZKW+19}. BLEU and chrF++ are calculated using SacreBLEU \cite{Pos18}.

\section{Evaluation}

Building on our experimental setup, this section presents the results of our multimodal translation evaluations. 


\subsection{Effect of Visual Grounding}

We begin by assessing translation quality and the effect of visual grounding using both human preference and automatic evaluations. 

\paragraph{Human Preferences Evaluation} Figure~\ref{fig:mmt-win-rates} shows native speaker preferences across 21 languages, comparing translations from text-only and text\textbf{+}image settings. We report win rates in instances where a preference was expressed between the two. Overall, translations with visual context are preferred above chance (50\%) in the majority of language–model combinations. Specifically, in the $X \to En$ direction, multimodal outputs are favored in 43 out of 63 experiments. A similar trend holds in the $En \to X$ direction, where text\textbf{+}image translations are preferred in 42 out of 63 cases. We observe that the text-only output was preferred in 37 out of the 126 total comparisons (29.4\%), while 4 out of 126 show a tie in preferences between modalities. These results suggest that visual grounding generally leads to translations that are more aligned with native speaker preferences, \textit{regardless of translation direction}.

\begin{figure*}[t!]
    \centering
    \includegraphics[width=\linewidth]{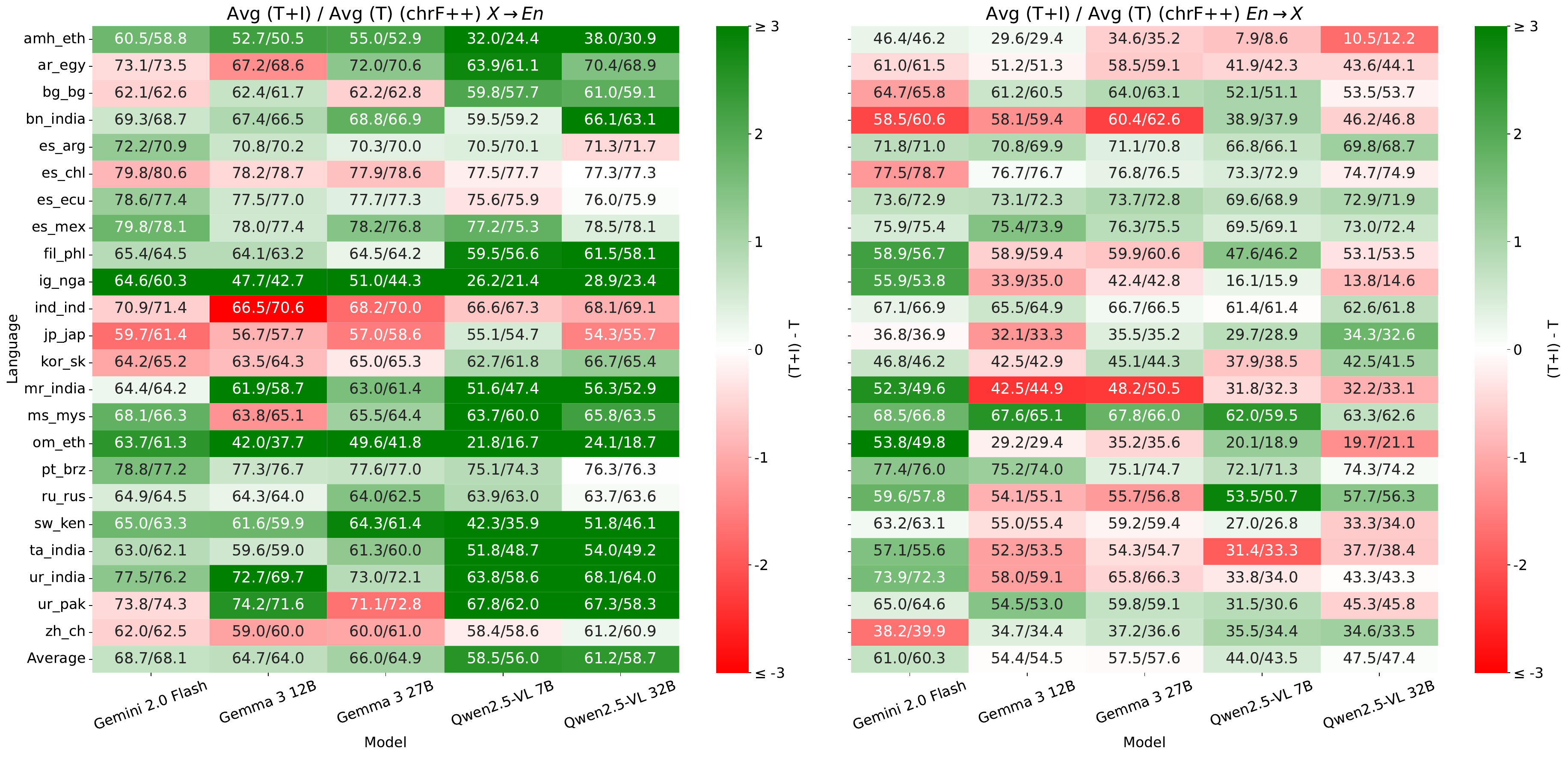}
    \caption{Heatmaps showing average chrF++ scores for text\textbf{+}image (T+I) and text-only (T) settings.
Left: Regional-to-English translation. Right: English-to-regional. Each cell shows (T+I) / (T) scores, with color indicating the difference, green shades represent 
improvements from image input.}
    \label{fig:chrf-comparison}
\end{figure*}

\paragraph{Automatic Evaluation} We base our main analyses on chrF++ as it has shown higher correlation with human judgments over BLEU \cite{popovic2017chrf++,Kocmi2021ship}. Figure~\ref{fig:chrf-comparison} reports chrF++ for 23 regions across 19 language pairs. In the $X\to En$ direction, most regions show improvements with image-grounded translations, with a few exceptions (e.g., Japan, Indonesia, and China). In the $En\to X$ direction, the benefit of multimodality is less consistent: while Gemini demonstrates clear gains, other models show mixed trends, with no systematic advantage or degradation from adding images. 
We present the results on BLEU and BertScore in Appendix \ref{sec:results_bleu}, which reflect a similar pattern. Additionally, in Appendix \ref{sec:chrf_categories} we report average chrF++ scores per CVQA-category.

\begin{table}[t]
\centering
\small
\resizebox{\columnwidth}{!}{%
\begin{tabular}{lllll}
\toprule
 &  & \multicolumn{1}{c}{$X\to En$} & \multicolumn{1}{c}{$En\to X$} \\
 \cmidrule(lr){3-3} \cmidrule(lr){4-4}
Model & Setting & chrF++ & chrF++ \\
\midrule
NLLB-600M & T & 56.9 & 50.3 \\
NLLB-3.3B & T & 58.9 & 54.9 \\
\midrule
Gemini 2.0 & T & 68.1 & 60.3 \\
Gemini 2.0 & T+I & 68.7 (\textbf{+0.7}) & 61.0 (\textbf{+0.7}) \\
\midrule
Gemma3 12B & T & 64.0 & 54.5 \\
Gemma3 12B & T+I & 64.7 (\textbf{+0.7}) & 54.4 (-0.1) \\
Gemma3 27B & T & 64.9 & 57.6 \\
Gemma3 27B & T+I & 66.0 (\textbf{+1.1}) & 57.5 (-0.1) \\
\midrule
Qwen2.5 VL 7B & T & 56.0 & 43.5 \\
Qwen2.5 VL 7B & T+I & 58.50 (\textbf{+2.5}) & 44.0 (\textbf{+0.5}) \\
Qwen2.5 VL 32B & T & 58.7 & 47.4 \\
Qwen2.5 VL 32B & T+I & 61.2 (\textbf{+2.5}) & 47.5 (\textbf{+0.1}) \\
\bottomrule
\end{tabular}%
}
\caption{
chrF++ scores averaged across languages for text-only (T) vs multimodal (T+I) settings in both directions ($X\to En$ and $En\to X$). The difference (T+I - T) is shown in parentheses.
}
\label{tab:model_chrf_comparison}
\end{table}

In Table~\ref{tab:model_chrf_comparison}, we report the average chrF++ scores across languages (for BLEU and BERT scores, refer to Appendix \ref{sec:results_bleu}). Notably, the addition of image context consistently improves performance across most VLMs, with gains most pronounced in the $X \to En$ direction.  Both evaluations support the conclusion that visual grounding improves translation quality for most languages, particularly when translating from regional languages to English. For the reverse direction, benefits are model-specific: native speakers still tend to prefer image-grounded translations from open-weight models, but this is not always reflected in automatic metrics. 

\begin{figure*}[t!]
    \centering
    \includegraphics[width=\linewidth]{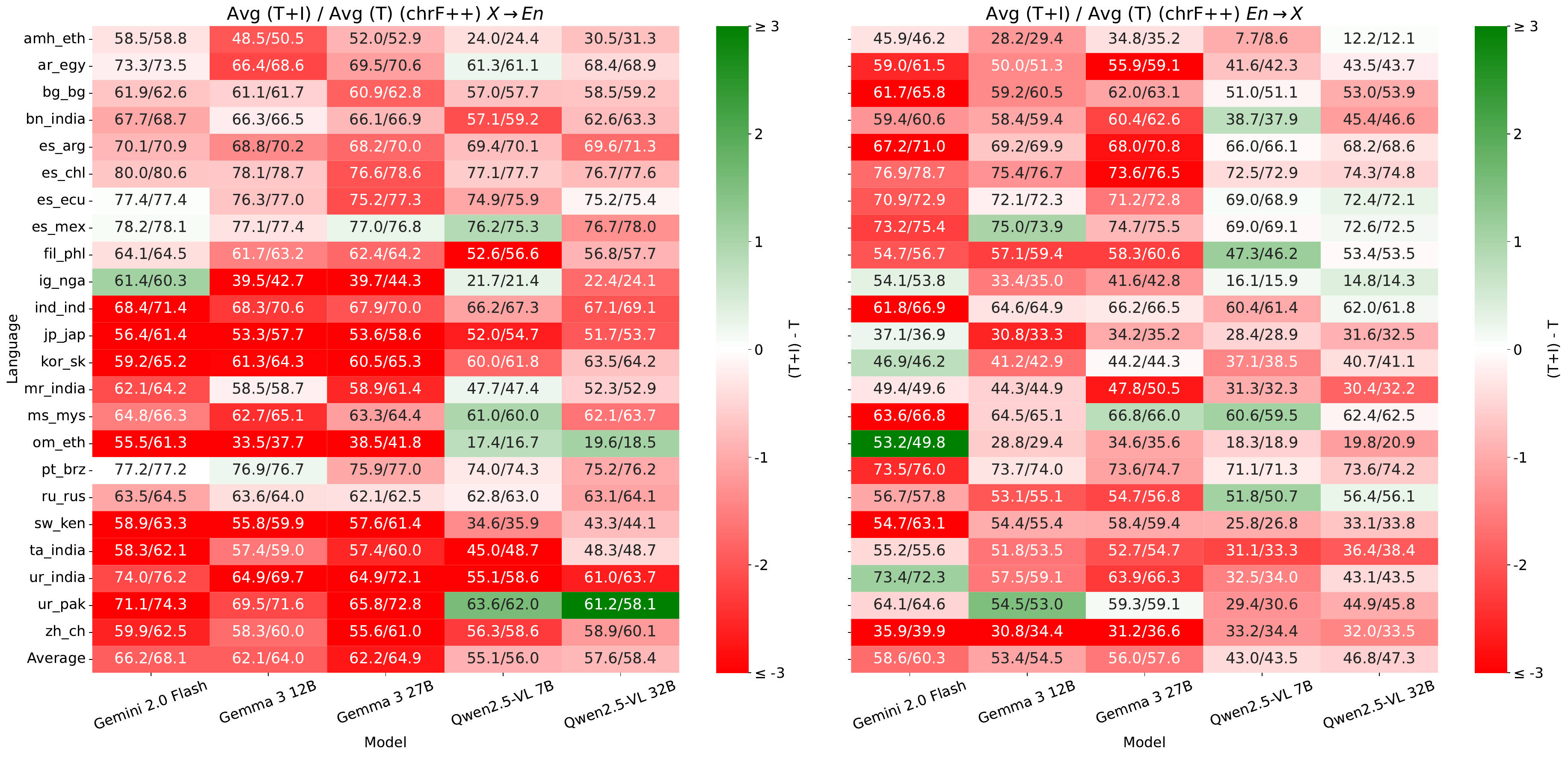}
    \caption{Average chrF++ scores for text+image (T+I) and text-only (T) settings when the image in the T+I setting is randomly sampled. Unlike Figure~\ref{fig:chrf-comparison}, where the image corresponds to the translated sentence, here we observe that unrelated images consistently lower chrF++ scores compared to the text-only setting.}
    \label{fig:shuffled}
\end{figure*}

To ensure that the observed gains are due to the images' content and not the fact that we are simply providing an image, we carry out a control experiment. We replicate the evaluation of this section using randomly sampled images (i.e., images from other items in \textsc{CaMMT}) in the T+I setting. We find that this consistently hurts performance, with chrF++ scores dropping across most languages and models (see Figure \ref{fig:shuffled}). Therefore, these results confirm that the observed gains are due to the actual image content rather than simply providing any image. At the same time, they show that unrelated images can be harmful to multimodal translation.

\subsection{Reasons Behind Preferences}

\begin{table*}[]
\centering
\small
\resizebox{1.8\columnwidth}{!}{%
\begin{tabular}{lllllllll}
\toprule
 & \multicolumn{2}{c}{$X\to En$} & \multicolumn{2}{c}{$En\to X$} & \multicolumn{2}{l}{$En_{Sub} \to X$} & \multicolumn{2}{l}{$En_{Cons} \to X$} \\
 & \# (T+I / T) & \%(T+I) & \# (T+I / T) & \%(T+I) & \# (T+I / T) & \%(T+I) & \# (T+I / T) & \%(T+I) \\
 \midrule
CSI-preserved & \textbf{380} / 277 & 57.8 & \textbf{304} / 203 & 60.0 & \textbf{223} / 147 & 60.3 & \textbf{139} / 79 & 63.8 \\
Gender & \textbf{33} / 2 & 94.3 & \textbf{45} / 36 & 55.6 & 10 / \textbf{12} & 45.5 & \textbf{10} / 6 & 62.5 \\
Disambiguation & \textbf{432} / 174 & 71.3 & \textbf{239} / 170 & 58.4 & \textbf{92} / 78 & 54.1 & \textbf{67} / 58 & 53.6 \\
Phrasing & \textbf{1329} / 1046 & 56.0 & \textbf{1238} / 1152 & 51.8 & \textbf{402} / 394 & 50.5 & \textbf{370} / 343 & 51.9 \\
Others & \textbf{368} / 320 & 53.5 & \textbf{301} / 289 & 51.0 & 56 / \textbf{74} & 43.1 & 86 / \textbf{98} & 46.7 \\
\bottomrule
\end{tabular}%
}
\caption{Breakdown of human preference reasons across translation directions. For each category, we report the number of times across all languages where a translation with image (T+I) or without image (T) was preferred, as well as the percentages for preferred (T+I). Numbers in bold indicate the modality with the highest preference. Results are shown for both directions and aggregated across languages and models.}
\label{tab:counts_preffs}
\end{table*}

To better understand how visual input influences translation decisions, we analyze the reasons provided by annotators during the human preference evaluation. Table~\ref{tab:counts_preffs} reports the number of preferences for text-only (T) versus text\textbf{+}image (T+I) translations, broken down by reason.

Across both directions, the primary factors driving preferences toward multimodal translation include \textit{CSI preservation}, \textit{correct gender}, and \textit{lexical disambiguation}. These effects are more pronounced in the $X\to En$ direction, where VLMs appear better at resolving gender and lexical ambiguities when images are available. The most common reason for preference is more regionally appropriate phrasing. While the T+I setting is still generally favored here, the margin over text-only is smaller, suggesting that visual input has a more modest impact on phrasing compared to other factors.

In preferences explained by annotators with \textit{other} reasons, which include reasons such as grammatical correctness, plural forms, and capitalization, the difference between T and T+I is also minimal, suggesting that images have a greater impact in resolving cultural or semantic ambiguity than in improving general linguistic quality.

We also examine native speakers' preferences in the \textit{conserved} and \textit{substituted} splits of \textsc{CaMMT} ($En_{Cons} \to X$ and $En_{Sub} \to X$), where the CSI in the source sentence has either been preserved or substituted. In these preferences (labeled with the CSI-preserved reason), speakers more frequently prefer translations from the T+I setting, implying that visual input helps models recover or preserve relevant cultural content. 

\paragraph{Models' Behavior on CSIs}
Beyond human preferences, we further analyze VLMs' ability to handle CSIs in translation. Specifically, we compute the average proportion of translations in which a CSI is preserved when the English source sentence contains a substituted ($En_{Sub} \to X$) or conserved ($En_{Cons} \to X$) CSI. 

In the \textit{substituted} setting, the source includes a substituted term instead of the CSI. %
We compute the percentage of times the model “recovers” the original CSI with the help of the image instead of keeping the substituted term or replacing it by other equivalent. In the \textit{conserved} setting, the source already contains the CSI, and we evaluate the percentage of times the model preserves it after translation. This analysis is independent of annotator preferences or ground-truth references, and instead probes how images bias models’ translation choices. As in prior experiments, models are not given explicit instructions about handling CSIs.

To do this, we use GPT-4o in a two-step process: (1) extract the CSI from the conserved version of each sample, and (2) check whether it appears in the model-generated translation. Details of this procedure are provided in Appendix~\ref{sec:alg_csi}. We then compute the percentage of CSI preservation by dividing the number of retained instances by the total number of samples.

Results presented in Table~\ref{tab:csis} show that, in the \textit{substituted} setting, the inclusion of images leads to a higher rate of CSI preservation, indicating the model's ability to retrieve appropriate region-specific concepts with visual grounding.  In the \textit{conserved} setting, the effect of images is less consistent: while CSIs are often preserved, we observe that image grounding can also lead to modifications of the terms, resulting in a small decrease in the proportion of retained CSIs compared to the text-only setting. These results suggest that images can bias models to recover CSIs that were previously replaced in English, but may also introduce variability in CSIs when the CSI is already conserved in the source.





\begin{table}[]
\centering
\small
\resizebox{0.85\columnwidth}{!}{%
\begin{tabular}{lllll}
\toprule
 & \multicolumn{2}{l}{$En_{Sub} \to X$} & \multicolumn{2}{l}{$En_{Cons} \to X$} \\
 \midrule
Model & T & T+I & T & T+I \\
\midrule
Qwen2.5 VL 32B & 20.27 & 23.05 & 80.70 & 77.83 \\
Gemma3 27B & 32.49 & 36.03 & 90.32 & 89.33 \\
Gemini 2.0 Flash & 41.72 & 44.05 & 91.24 & 90.91 \\
\bottomrule
\end{tabular}%
}
\caption{Average percentage of preserved CSIs across languages. A value of 100 indicates that all CSIs are retained in the translation; 0 indicates none are preserved. Appendix \ref{sec:alg_csi} reports per-language differences and the average impact of images.}

\label{tab:csis}
\end{table}

\begin{table}[]
\centering
\small
\resizebox{\columnwidth}{!}{%
\begin{tabular}{ccccc}
\toprule
                             & Forced-C    & Forced-S    & Possible-C   & Possible-S  \\
\midrule
Latin      & 94\(\pm\)6.7     & 6\(\pm\)6.7      & 63\(\pm\)29.4     & 37\(\pm\)29.4    \\
Non-Latin  & 75\(\pm\)36.2    & 25\(\pm\)36.2    & 53\(\pm\)20.2     & 47\(\pm\)20.2    \\
\bottomrule
\end{tabular}%
}
\caption{Translation preferences when curating \textsc{CaMMT}. 
Annotators classified each CSI as either having a `Forced' translation or having a `Possible' translation. 
`C' and `S' represent \textit{conserved} and \textit{substituted} translations, respectively.}
\label{tab:translation_comparison}
\end{table}

\subsection{Comparisons of VLMs' Performance}
We assess the overall MT performance of VLMs. Firstly, as shown in Figure \ref{fig:chrf-comparison} and Table~\ref{tab:model_chrf_comparison}, the best performance is achieved by Gemini, followed by Gemma and Qwen models. 
Secondly, we compare their performance against a strong text-based MT baseline. As shown in Table~\ref{tab:model_chrf_comparison}, 
compared to NLLB-3.3B, the best-performing VLMs (Gemini 2.0 and Gemma3-27B) achieve comparable or superior translation performance in most metrics, particularly in the $X \to En$ direction, where they show considerable advantages.

\subsection{Human Translation Preferences for CSIs}

This section examines native speakers' 
preferred translation strategies when handling CSIs at the moment of curating \textsc{CaMMT}, where we examine their patterns across languages with different script types.  Table \ref{tab:translation_comparison} presents the percentage distribution of human preferences for \textit{conserved} versus \textit{substituted} translations for Latin and non-Latin scripts under two distinct conditions: when the CSI has a similar equivalent in English (\textit{conserved}), against the case in which there is no equivalent (\textit{forced)}.

For forced translations, annotators with Latin script languages strongly favored conservation. Annotators with non-Latin scripts also leaned towards conservation, but were more open to substitution. When possible translations existed, both script types demonstrated a more balanced choice between the two strategies. 

\section{Discussion}
In this section, we revisit our research questions in light of the experimental findings.
\paragraph{\colorbox{rq1}{RQ1} \& \colorbox{rq2}{RQ2}: What is the impact of visual grounding on translation quality, and what factors explain this effect?}
Visual grounding generally improves translation quality, particularly in ways that are meaningful to human evaluators. While gains in automatic metrics such as BLEU and chrF++ may appear modest, human preference evaluations tell a richer story: In 85 out of 126 model–language–direction comparisons (67.5\%) where a preference was stated, native speakers preferred multimodal translations, underscoring the value of images for improving cultural and semantic alignment of translations. 

Reasons for preference, shown in Table \ref{tab:counts_preffs}, reveal that images are particularly helpful in preserving CSIs, correcting gender, and improving disambiguation. These improvements often involve small textual changes that can significantly impact perceived quality, but may not strongly affect automatic metrics. We conclude that, \textbf{visual grounding seems to strengthen translation quality primarily by supporting semantic precision and cultural retention}, benefits that are better captured by human judgments than by traditional MT metrics.

That said, in 37 out of 126 comparisons (29.4\%), text-only translations were preferred, indicating that visual input can occasionally degrade translation quality. Understanding why this occurs remains an open question and is an important direction for future work. Moreover, the relatively small gains in automatic metrics are consistent with patterns observed in earlier multimodal MT studies \cite{futeral2022tackling}, underscoring the need for improved evaluation methods that more accurately reflect the contribution of visual context, particularly in multicultural scenarios. Finally, we observed that unrelated images can negatively affect translation; therefore, future work should also study how to develop new models that can, on the fly, decide when visual context should influence translation to improve the robustness of these systems to noisy visual information.

\paragraph{\colorbox{rq3}{RQ3}: 
How do VLMs perform in MT compared to each other and to specialized systems?
}
In terms of Machine Translation performance, all evaluated VLMs matched or exceeded the performance of strong baselines like NLLB-600M and 3.3B, where the closed-source model (Gemini 2.0 Flash) outperformed open-weight models (Qwen2.5 and Gemma3 families). Notably, \textbf{we do not observe an evident tradeoff when using VLMs for translation}: they offer competitive performance in standard metrics while simultaneously providing the ability to leverage visual context. This highlights their potential as general-purpose translation systems capable of steering translations using multimodal inputs without sacrificing textual quality.

\paragraph{\colorbox{rq4}{RQ4}: 
Which translation strategies do native speakers prefer in the case of CSIs?
}

Contrary to the predominant research direction in NLP on substitution strategies for unfamiliar CSIs, our findings suggest that native speakers often prefer conservation, especially when no culturally equivalent term exists in English. This trend holds across both Latin and non-Latin scripts, although the latter group shows greater variability.
When equivalents are available, preferences are more balanced, but still do not lean completely toward substitution. These results point to the importance of incorporating script-aware translation strategies regarding CSIs in future research, highlighting the need for MT systems to better align with native speaker preferences by adapting conservation and substitution choices to regional and linguistic contexts.

\section{Conclusions}
We present \textsc{CaMMT}, a human-curated dataset for Multimodal Machine Translation that encompasses 19 languages across 23 regions. We evaluated five VLMs at different scales on \textsc{CaMMT} and observed that providing images as auxiliary context generally improves translation quality in ways that native speakers find meaningful. When translations incorporate images, they tend to better preserve cultural elements, use correct gender marking, and resolve ambiguities. All of these improvements are often overlooked by automatic MT metrics. However, we also observe a non-trivial number of cases where visual input negatively affects translations. Understanding when and why this occurs remains an important direction for future research.

Our findings also show that annotators tend to favor conserving CSIs, particularly when no clear equivalent exists in English, underscoring the importance of culturally sensitive translation strategies. Future work should incorporate such speaker-aligned choices when designing models and datasets for grounded, culturally aware translation.

\section*{Limitations}
While \textsc{CaMMT} provides broad language and regional coverage, the number of samples is constrained by the original CVQA dataset. Due to design choices inherited from CVQA, some samples are marked as non-culturally relevant; however, we retain them as they remain useful for evaluating general multimodal machine translation. 
When curating \textsc{CaMMT}, we relied on a single annotator per region for human annotations, which may introduce subjectivity in CSI assessments and translation preferences. Expanding annotator diversity would likely improve the reliability and objectivity of these judgments. On the evaluation side, we do not evaluate specialized MMT systems, as most lack training data for the 19 languages included. To keep human evaluation feasible across three models, we restrict evaluation to pairwise preferences between text-only and text\textbf{+}image outputs. We do not include Likert-scale judgments of translation quality, relying primarily on automatic metrics for this purpose. Future work should explore how visual grounding affects human perception of translation quality, as well as expand the dataset with more samples per region and involve multiple annotators to improve coverage and objectiveness of cultural relevance and CSI judgments.

\bibliography{custom}

\appendix
\section{Appendix}
\subsection{\textsc{CaMMT} Statistics}
\label{sec:data_statistics}
In Table \ref{tab:dataset_stat}, we report the number of samples per region in \textsc{CaMMT}, their language and writing script. In addition, we include number of samples that are: CSIs (Forced translation or Has possible translations), Culturally Relevant (non-CSI), or Not culturally relevant.

\begin{table*}[b]
\centering
\resizebox{0.75\textwidth}{!}{%
\begin{tabular}{lllcccc}
\toprule
\textbf{Language-Region} & \textbf{Script(s)} & \textbf{Size} & \multicolumn{2}{c}{\textbf{CSI}} &\makecell{\textbf{Culturally Relevant}\\\textbf{(non-CSI)}}
 & \makecell{\textbf{Not culturally}\\\textbf{relevant}} \\
\cmidrule(lr){4-5}
& & & \textbf{Forced} & \textbf{Possible} & & \\
\midrule
Amharic-Ethiopia & Ge'ez & 234 & 31 & 49 & 97 & 57 \\
Arabic-Egypt & Arabic & 203 & 16 & 8 & 95 & 84 \\
Bengali-India & Bengali & 286 & 54 & 31 & 61 & 140 \\
Bulgarian-Bulgaria & Cyrillic & 369 & 8 & 19 & 90 & 252 \\
Chinese-China & Hanzi & 308 & 26 & 18 & 152 & 112\\
Filipino-Philippines & Latin (Rumi) & 203 & 26 & 29 & 20 & 128 \\
Igbo-Nigeria & Latin & 200 & 22 & 41 & 62 & 75\\
Indonesian-Indonesia & Latin (Rumi) & 202 & 29 & 7 & 81 & 85\\
Japanese-Japan & Kanji & 203 & 46 & 26 & 51 & 80\\
Korean-South Korea & Hangul & 290 & 51 & 11 & 103 & 125 \\
Malay-Malaysia & Latin (Rumi) & 315 & 48 & 40 & 196 & 31\\
Marathi-India & Devanagari & 202 & 27 & 25 & 99 & 51 \\
Oromo-Ethiopia & Latin & 214 & 51 & 70 & 93 & 0 \\
Portuguese-Brazil & Latin & 284 & 46 & 31 & 203 & 4 \\
Russian-Russia & Cyrillic & 200 & 31 & 26 & 31 & 112 \\
Spanish-Argentina & Latin & 265 & 32 & 50 & 55 & 128\\
Spanish-Chile & Latin & 234 & 34 & 49 & 73 & 78 \\
Spanish-Ecuador & Latin & 362 & 12 & 60 & 70 & 220\\
Spanish-Mexico & Latin & 323 & 12 & 67 & 94 & 150 \\
Swahili-Kenya & Latin & 271 & 43 & 99 & 124 & 5 \\
Tamil-India & Tamil & 213 & 32 & 16 & 44 & 121\\
Urdu-India & Perso-Arabic & 220 & 27 & 22 & 97 & 74\\
Urdu-Pakistan & Perso-Arabic & 216 & 24 & 28 & 120 & 44\\
\bottomrule
\end{tabular}%
}
\caption{\textbf{Languages covered in \textsc{CaMMT} and Dataset statistics}: including writing script, region, number of samples, and CI counts. Each region was annotated by native speaker.}
\label{tab:dataset_stat}
\end{table*}

We use statistics of this dataset (specifically, scripts of each language), to understand translations choices of annotators when it comes to conserving or substituting CSIs.

\subsection{Experimental Setting}
\label{sec:experimental_setting}
We employ the \textit{transformers} library \cite{wolf2020huggingfacestransformersstateoftheartnatural} for all the experiments conducted on open-weight models. The specific identifiers for each model are shown in Table \ref{tab:models}. All experiments are run on single NVIDIA A100 80G card. We set temperature to 0.0 for generating the translations.

Following \citet{cavalin2025sentence}, we evaluate chrF++ and BLEU scores at sentence-level using SacreBLEU \cite{post-2018-call}. BERTScore is calculated using \textit{bert-base-multilingual-cased} model for all languages\footnote[1]{https://github.com/Tiiiger/bert\_score}  at corpus-level.

\subsection{Aya-Vision Evaluations}
\label{sec:aya_appendix}

\begin{figure*}[t!]
    \centering
    \includegraphics[width=\linewidth]{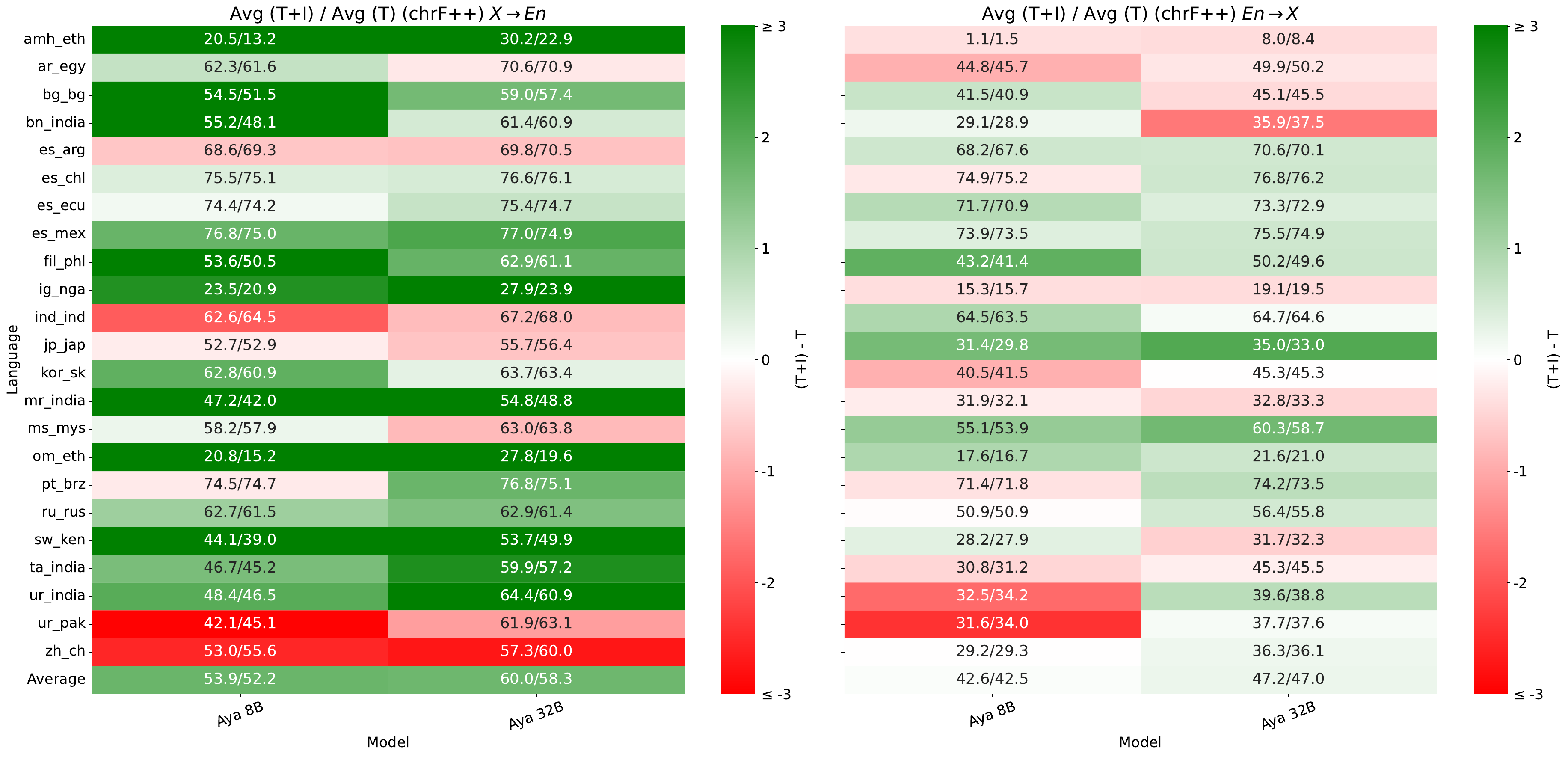}
    \caption{Heatmaps showing average chrF++ scores in the Aya-Vision family for text\textbf{+}image (T+I) and text-only (T) settings. We observe a consistent behavior with the models evaluated in our main analysis.}
    \label{fig:chrf-comparison-aya}
\end{figure*}

In Figure~\ref{fig:chrf-comparison-aya}, we report chrF++ scores for Aya-Vision 8B and 32B. The results show a pattern consistent with Gemini, Gemma 3, and Qwen2.5-VL. In the $X \to$ En direction, providing the image generally improves performance, yielding higher scores in most cases. In the En $\to X$ direction, the effect is more mixed, though it remains mostly positive, particularly for the larger model. Overall, these results complement and reinforce our main findings.

\begin{table}[H]
\small
\centering
\resizebox{0.9\columnwidth}{!}{%
\begin{tabular}{ll}
\toprule
\textbf{Model} & \textbf{Hugging Face Identifier} \\
\midrule
Gemma 3 27B\footnotemark[2] & \texttt{google/gemma-3-27b-it} \\
Gemma 3 12B\footnotemark[3] & \texttt{google/gemma-3-12b-it} \\
Qwen2.5-VL 32B\footnotemark[4] & \texttt{Qwen/Qwen2.5-VL-32B-Instruct} \\
Qwen2.5-VL 7B\footnotemark[5] & \texttt{Qwen/Qwen2.5-VL-7B-Instruct} \\
AyaVision 32B\footnotemark[6] & \texttt{CohereForAI/aya-vision-32b} \\
AyaVision 8B\footnotemark[7] & \texttt{CohereForAI/aya-vision-8b} \\
\bottomrule
\end{tabular}%
}
\caption{HuggingFace identifiers for models used in our experiments.}
\label{tab:models}
\end{table}

\footnotetext[2]{https://huggingface.co/google/gemma-3-27b-it}
\footnotetext[3]{https://huggingface.co/google/gemma-3-12b-it}
\footnotetext[4]{https://huggingface.co/Qwen/Qwen2.5-VL-32B-Instruct}
\footnotetext[5]{https://huggingface.co/Qwen/Qwen2.5-VL-7B-Instruct}
\footnotetext[6]{https://huggingface.co/CohereForAI/aya-vision-32b}
\footnotetext[7]{https://huggingface.co/CohereForAI/aya-vision-8b}

\subsection{BLEU and BertScore metrics across models}
\label{sec:results_bleu}
\begin{figure*}[htpb]
    \centering
    
    \begin{subfigure}{\textwidth}
        \centering
        \resizebox{\textwidth}{!}{%
            \includegraphics{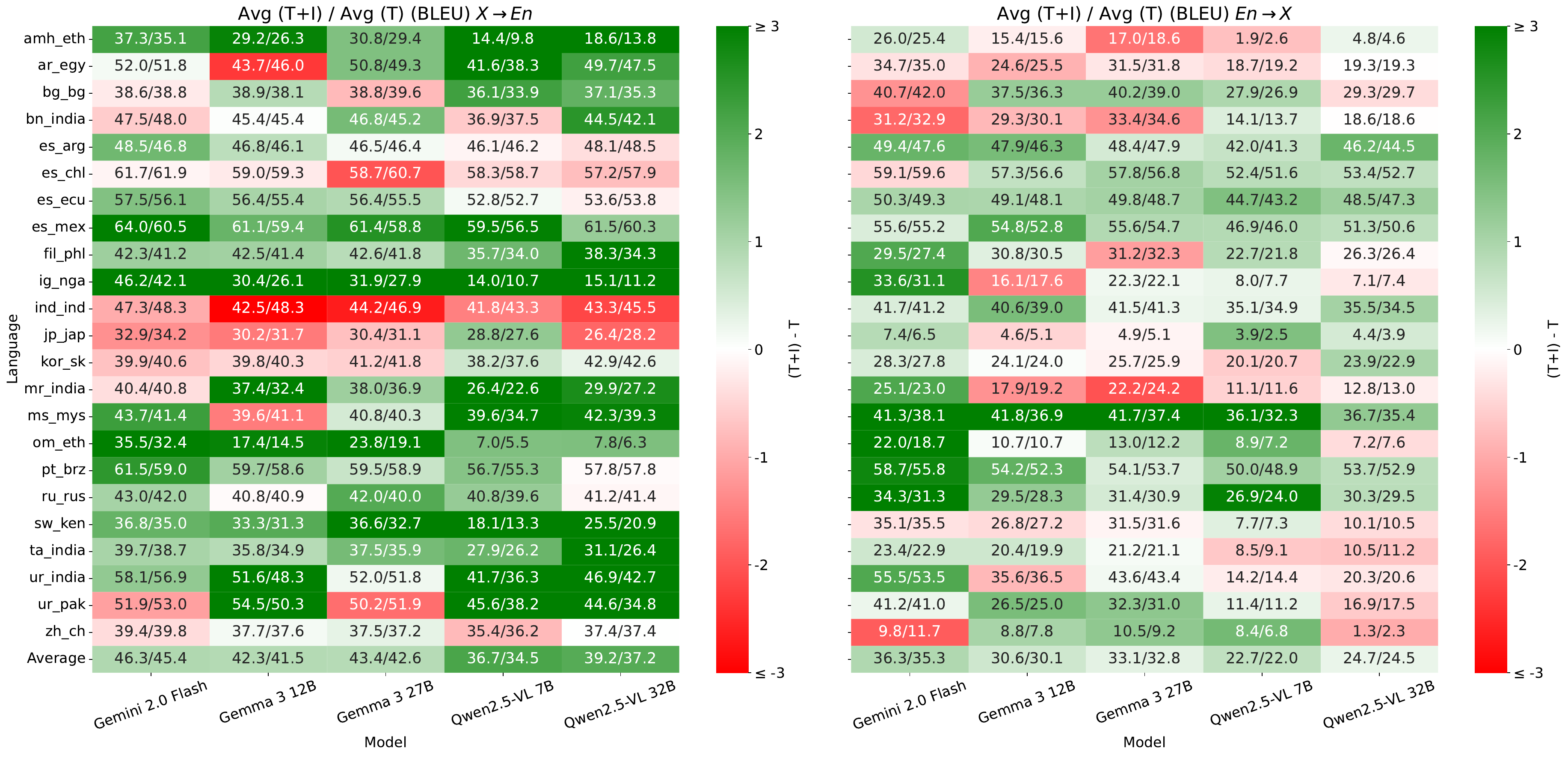}
        }
        \caption{BLEU scores comparison}
        \label{fig:bleu-comparison}
    \end{subfigure}
    
    \vspace{0.4cm} 
    
    \begin{subfigure}{\textwidth}
        \centering
        \resizebox{\textwidth}{!}{%
            \includegraphics{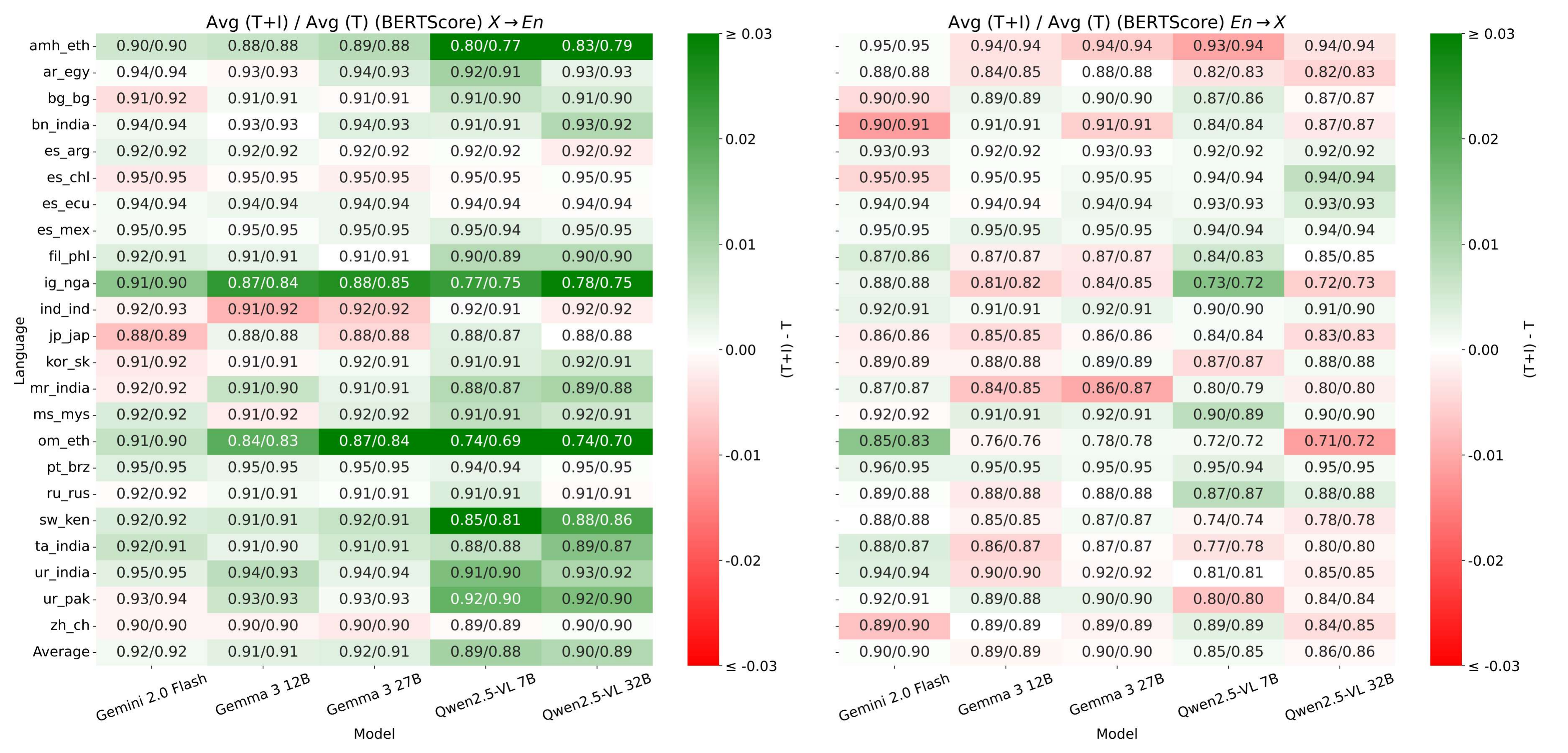}
        }
        \caption{BERT scores comparison}
        \label{fig:bert-comparison}
    \end{subfigure}
    
    \caption{Heatmaps showing the difference in average BLEU and BERT scores for text\textbf{+}image (T+I) and text-only (T) settings. 
    Left: Regional-to-English translation. Right: English-to-regional. Each cell shows (T+I) / (T) scores, with color indicating the difference, green shades represent improvements from image input.
}
    \label{fig:score-comparison}
\end{figure*}

In Table~\ref{tab:bleu_comparison}, we calculate BLEU and BERTScore metrics for both MMT and text-based translations averaged across languages for all models. We also present heatmaps in Figure~\ref{fig:score-comparison} showing the results for each language, providing a comparison between the performance of MMT and text-based settings.

\begin{table}[!htbp]
\centering
\resizebox{0.95\columnwidth}{!}{%
\begin{tabular}{lllllll}
\toprule
 &  & \multicolumn{2}{c}{$En\to X$} & \multicolumn{2}{c}{$X\to En$} \\
 \cmidrule(lr){3-4} \cmidrule(lr){5-6}
Model & Setting & BLEU & BERT & BLEU & BERT \\
\midrule
NLLB-3.3B & T & 28.98 &  & 36.01 &  \\
\midrule
Gemini 2.0 & T & 35.70 & 0.9 & 45.60 & 0.92 \\
Gemini 2.0 & T+I & 36.56 (\textbf{+0.87}) & 0.9 & 46.51 (\textbf{+0.91}) & 0.92 \\
\midrule
Gemma3 12B & T & 30.03 & 0.89 & 41.46 & 0.91 \\
Gemma3 12B & T+I & 30.62 (\textbf{+0.59}) & 0.89 & 42.33 (\textbf{+0.87}) & 0.91 \\
Gemma3 27B & T & 32.78 & 0.9 & 42.60 & 0.91 \\
Gemma3 27B & T+I & 33.17 (\textbf{+0.38}) & 0.9 & 43.45 (\textbf{+0.85}) & 0.92 (\textbf{+0.01}) \\
\midrule
Qwen VL 7B & T & 21.96 & 0.85 & 34.57 & 0.88 \\
Qwen VL 7B & T+I & 22.68 (\textbf{+0.72}) & 0.85 & 36.71 (\textbf{+2.14}) & 0.89 (\textbf{+0.01})  \\
Qwen VL 32B & T & 24.39 & 0.86 & 37.32 & 0.89 \\
Qwen VL 32B & T+I & 24.65 (\textbf{+0.26}) & 0.86 & 39.21 (\textbf{+1.89}) & 0.90 (\textbf{+0.01}) \\
\bottomrule
\end{tabular}%
}
\caption{
BLEU and BERT scores averaged across languages for text-only (T) vs multimodal (T+I) settings in both directions ($En\to X$ and $X\to En$). The difference (T+I - T) is shown in parentheses.
}
\label{tab:bleu_comparison}
\end{table}

\subsection{Translation Prompts}
\label{sec:translation_prompts}

In our experiments, we use two types of prompts for translation tasks: text-only translation (MT) and multimodal translation (MMT). The prompts are defined as follows:

\begin{lstlisting}[basicstyle=\ttfamily\small, breaklines=true, frame=single]
PROMPT_MT = '''Translate the following sentence from {source} to {target}. Provide ONLY the translated text, with no additional information, explanation, or context.
"{sentence}"
'''
\end{lstlisting}

\begin{lstlisting}[basicstyle=\ttfamily\small, breaklines=true, frame=single]
PROMPT_MMT = '''Translate the following sentence from {source} to {target} using the provided image as additional context. Provide ONLY the translated text, with no additional information, explanation, or context.
"{sentence}"
'''
\end{lstlisting}

Where \texttt{PROMPT\_MT} was used for text-only translation (T) and \texttt{PROMPT\_MMT} was used for multimodal translation with text and image (T+I).

\subsection{Comparison between categories measured by chrF scores}
\label{sec:chrf_categories}

\begin{figure*}[ht!]
    \centering
    \resizebox{\textwidth}{!}{%
        \includegraphics{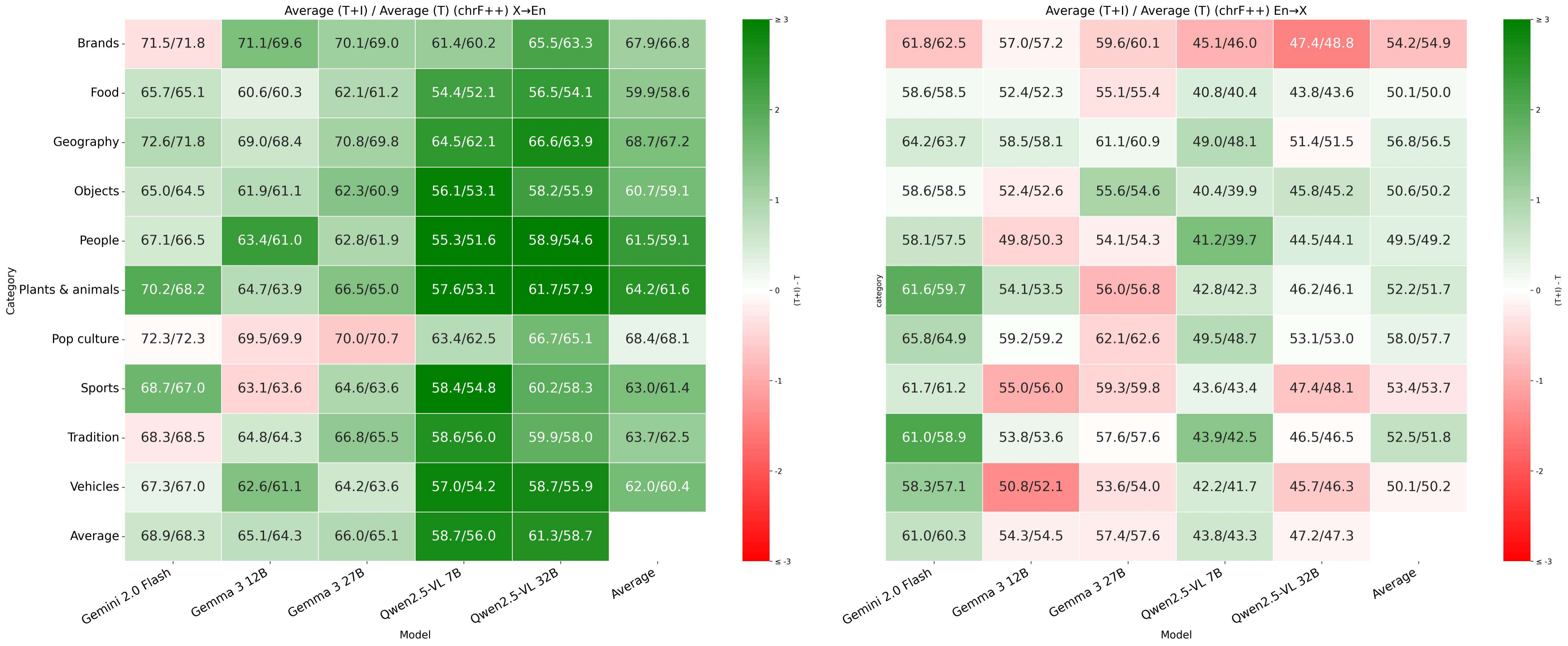}
     }
    \caption{
    Heatmaps showing the difference in average chrF++ scores for text\textbf{+}image (T+I) and text-only (T) across categories and models. Left: Regional-to-English translation. Right: English-to-regional. Each cell shows (T+I) / (T) scores, with color indicating the difference, green shades represent improvements from image input.
    }
    \label{fig:cat-comparison}
\end{figure*}

The original CVQA dataset encompasses questions across 10 diverse categories: vehicles, food, people, sports, plants \& animals, objects, brands, geography, tradition, and pop culture. Figure~\ref{fig:cat-comparison} shows automatic evaluation using CHRF++ scores across models and CVQA categories. 

In the $En \to X$ direction, the impact of visual input is notably selective. Only the \textit{geography} and \textit{traditions} categories consistently benefit from multimodal input across all models.
The $X \to En$ direction presents a 
different pattern, where visual context provides substantial benefits across most categories. Interestingly, two categories consistently show minimal benefits from visual input in $X \to En$ direction: \textit{brands} and \textit{pop culture}.

\subsection{License}

CVQA \cite{romero2024cvqa} allows using their QA data for research purposes, which is the aim of this work. We do not include the images in our release, and instead include their ID in CVQA. Refer to \citet{romero2024cvqa} for the licenses of the images, as each has a specific license.

The \textsc{CaMMT} corpus is exclusively for academic research, under the Creative Commons Attribution-NonCommercialShareAlike 4.0 International (CC BY-NC-SA 4.0) license. 

\clearpage
\onecolumn
\subsection{\textsc{CaMMT} Data Curation Guideline} 
\label{sec:annotation_guideline}
\begin{figure}[H]  
  \centering
  \includegraphics[width=\linewidth]{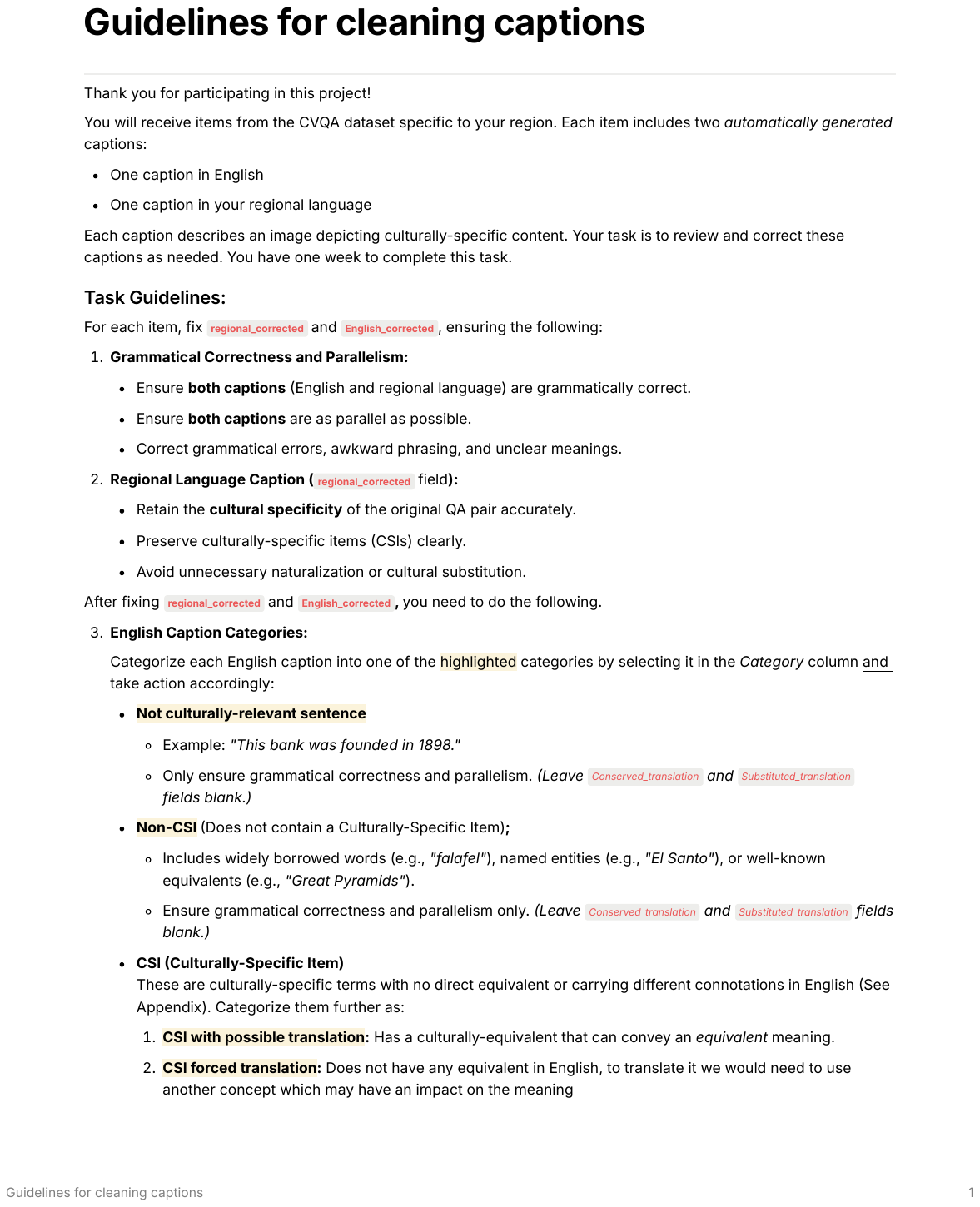}
  \caption{Annotation guideline}
  \label{fig:singlepdf}
\end{figure}
\twocolumn
\subsection{Human Preference Evaluation Instructions} \label{sec:preference_guidelines}

\small
\begin{quote}
\textbf{Instructions for Translation Evaluation Task}

You are tasked with selecting your preferences on the provided evaluation sheet. Each item includes:
\begin{itemize}
    \item A \textbf{source sentence}
    \item Two \textbf{model translations} (Model A and Model B)
    \item The \textbf{target translation} you previously created
    \item A \textbf{reference image} to help you disambiguate or contextualize cultural elements
\end{itemize}

Please fill the following columns:
\begin{enumerate}
    \item \textbf{Translation Quality}:
    \begin{itemize}
        \item Indicate whether one translation is better, both are good, or both are bad/unintelligible.
    \end{itemize}
    
    \item \textbf{Translation Preference}:
    \begin{itemize}
        \item Choose \textbf{A} or \textbf{B} based on which translation you prefer.
        \item Try to select one even if both are equally good or bad.
    \end{itemize}
    
    \item \textbf{Reason for Preference}:
    \begin{itemize}
        \item If you selected one translation as better, choose a reason from the predefined list.
        \item If no reason applies, explain briefly in the ``Other Reasons'' column (a few words are enough).
    \end{itemize}
    
    \item In the case of `both are good':
    \begin{itemize}
        \item \textbf{If both translations are essentially identical and equally good (e.g., differing only in word order), you may leave the preference entry blank.}
    \end{itemize}
\end{enumerate}
\end{quote}

\subsection{CSI Retention Evaluation}
\label{sec:alg_csi}

\begin{figure*}[!htbp]
\centering
\includegraphics[width=\textwidth]{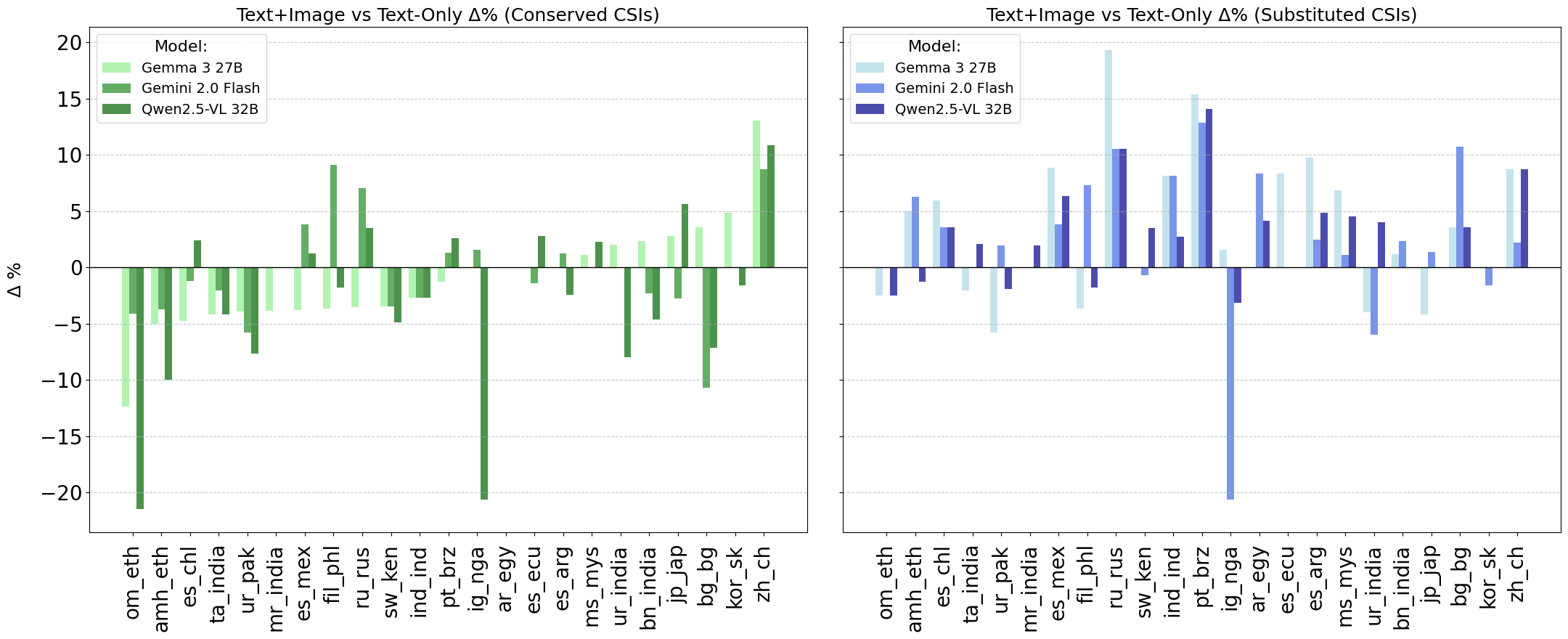}
\caption{Differences in CSI retention percentages between text-only and text\textbf{+}image settings for Gemma 3 (27B), Gemini 2.0 Flash, and Qwen2.5-VL 32B across languages. Left: conserved CSIs; right: substituted CSIs.}
\label{fig:csi-retention-score-comparison}
\end{figure*}

In this section, we report the per-language analysis of the impact of visual input on the retention of CSIs across languages (comparing text-only and text\textbf{+}image settings) and describe the algorithm for CSI identification in translations.

For each language and model, we compute the difference in CSI preservation rates using translations from the conserved and substituted splits. As shown in Table~\ref{tab:csis}, and further illustrated in Figure~\ref{fig:score-comparison}, visual input tends to help models recover CSIs in the substituted setting—where the original term is not present in the source sentence, by providing complementary visual cues. In contrast, when translating from the conserved split, where the CSI is explicitly present in the source, we observe no consistent effect from the image across models or languages. 

\vspace{5pt}

\paragraph{CSI extraction and identification} We developed a two-stage approach to evaluate how well machine translation systems preserve CSIs. This methodology leverages large language models to first identify CSIs and then evaluate their preservation in different translation outputs.

Our methodology consists of two key stages:
\begin{enumerate}
        \item \textbf{CSI Extraction}: Automatically identifying CSI using the prompt shown in Box~\ref{box:csi-extraction}, which compares conserved translations (containing the CSI) against substituted translations (where the CSI is replaced with a more general term).
    \item \textbf{CSI Preservation Evaluation}: Determining which of two competing translation systems better preserves the identified CSI when compared to a gold reference, following the evaluation setup in Box~\ref{box:compare}.
\end{enumerate}

For both CSI extraction and evaluation, we utilized GPT-4o with $temperature=0.0$ to ensure deterministic outputs. The CSI extraction was limited to $max\_tokens=50$, while we used default token limits for the evaluation task. All processing was performed through the OpenAI API, maintaining consistent parameters across all language pairs and translation systems.

\onecolumn
\clearpage
\small
\begin{tcolorbox}[
    colback=green!5!white,
    colframe=green!50!black,
    title=CSI Extraction Prompt,
    label={box:csi-extraction},
    width=\textwidth,  
    before skip=0pt,   
    after skip=10pt    
]
\textbf{Given two versions of a sentence:}
\begin{enumerate}
\item A sentence with a culturally specific item (conserved\_translation)
\item A sentence where that item has been replaced with a more general term (substituted\_translation)
\end{enumerate}
\textbf{Your task is to identify the culturally specific item (CSI)} that appears only in the conserved translation.

Compare the two sentences and extract only the specific culturally-significant word or phrase that was replaced in the substituted version.

\textbf{Return ONLY the culturally specific item} as a single word or phrase, without any explanations, quotation marks, or additional text.

\textbf{Example:}

Conserved: "The person in the picture is a famous \textbf{charro} from the state of Jalisco."

Substituted: "The person in the picture is a famous \textbf{cowboy} from the state of Jalisco."

Output: \textbf{charro}

\vspace{0.5em}
\ldots
\label{box:csi-extraction}
\end{tcolorbox}

\begin{tcolorbox}[
    colback=green!5!white,
    colframe=green!50!black,
    title=CSI Evaluation Prompt,
    label={box:compare},
    width=\textwidth,  
    before skip=0pt,   
    after skip=10pt    
]
\textbf{Given two translations (0 and 1), a gold reference sentence (y), and a culturally specific item (CSI)}, your task is to:

\textbf{Evaluate which translation better preserves the CSI from the reference.}

Output the results strictly as a JSON list of dictionaries with the following exact structure:
\begin{verbatim}
[
  {
    "word": [word_in_0, word_in_1, word_in_y],
    "type": "CSI",
    "aligned_translation": "0" | "1" | "None" | "both"
  }
]
\end{verbatim}

\textbf{Where "aligned\_translation" values mean:}
\begin{itemize}
\item \textbf{"0"}: Translation 0 better preserves the CSI
\item \textbf{"1"}: Translation 1 better preserves the CSI
\item \textbf{"both"}: Both translations include the provided CSI
\item \textbf{"None"}: None of the translations includes the original CSI (it is replaced by another term)
\end{itemize}

\textbf{Example 1:}

\textbf{Input:}\\
y: Este personaje es un \textbf{charro} famoso\\
0: Este personaje es un \textbf{vaquero} famoso\\
1: Este personaje es un \textbf{charro} famoso\\
csi: \textbf{charro}

\textbf{Output:}\\
\begin{verbatim}
[{"word": ["vaquero", "charro", "charro"], "type": "CSI", "aligned_translation": "1"}]
\end{verbatim}

\vspace{0.5em}
\ldots
\end{tcolorbox}
\onecolumn
\clearpage
\twocolumn
\subsection{Affiliations}
This section outlines the affiliations of each of the co-authors of this work:
\begin{itemize}
    \item Emilio Villa-Cueva (MBZUAI),
    \item Sholpan Bolatzhanova (MBZUAI),
    \item Diana Turmakhan (MBZUAI),
    \item Kareem Elzeky (MBZUAI),
    \item Henok Biadglign Ademtew (Vella AI),
    \item Alham Fikri Aji (MBZUAI),
    \item Vladimir Araujo (Sailplane AI),
    \item Israel Abebe Azime (Saarland University),
    \item Jinheon Baek (KAIST),
    \item Frederico Belcavello (Federal University of Juiz de Fora, CNPq),
    \item Fermin Cristobal (MBZUAI),
    \item Jan Christian Blaise Cruz (MBZUAI),
    \item Mary Dabre (Independent Researcher),
    \item Raj Dabre (IIT Madras),
    \item Toqeer Ehsan (Independent Researcher),
    \item Naome A Etori (University of Minnesota -Twin Cities),
    \item Fauzan Farooqui (MBZUAI),
    \item Jiahui Geng (MBZUAI),
    \item Guido Ivetta (Universidad Nacional de Córdoba, Argentina),
    \item Thanmay Jayakumar (IIT Madras),
    \item Soyeong Jeong (KAIST),
    \item Zheng Wei Lim (The University of Melbourne),
    \item Aishik Mandal (Technische Universität Darmstadt),
    \item Sofía Martinelli (Universidad Nacional de Córdoba, Argentina),
    \item Mihail Minkov Mihaylov (MBZUAI),
    \item Daniil Orel (MBZUAI),
    \item Aniket Pramanick (Technische Universität Darmstadt),
    \item Sukannya Purkayastha (Technische Universität Darmstadt),
    \item Israfel Salazar (University of Copenhagen),
    \item Haiyue Song (NICT),
    \item Tiago Timponi Torrent (Federal University of Juiz de Fora, CNPq),
    \item Debela Desalegn Yadeta (Addis Ababa University),
    \item Injy Hamed (MBZUAI),
    \item Atnafu Lambebo Tonja (MBZUAI),
    \item Thamar Solorio (MBZUAI)
\end{itemize}

\end{document}